\documentclass[10pt,twocolumn,letterpaper]{article}

\usepackage[final]{cvpr}
\usepackage{times}
\usepackage{epsfig}
\usepackage{graphicx}
\usepackage{amsmath}
\usepackage{amssymb}

\usepackage{subcaption}
\usepackage{todonotes}
\usepackage{multirow}
\usepackage{booktabs}
\usepackage{comment}

\usepackage{pifont}
\newcommand{\cmark}{\ding{51}}%
\newcommand{\xmark}{\ding{55}}%

\usepackage{amsmath,amsfonts,bm}

\def\eqref#1{equation~\ref{#1}}

\def\1{\bm{1}}

\def\vx{{\bm{x}}}

\def\mI{{\bm{I}}}

\DeclareMathAlphabet{\mathsfit}{\encodingdefault}{\sfdefault}{m}{sl}
\SetMathAlphabet{\mathsfit}{bold}{\encodingdefault}{\sfdefault}{bx}{n}

\DeclareMathOperator*{\argmin}{arg\,min}

\usepackage[pagebackref=true,breaklinks=false,letterpaper=true,colorlinks,bookmarks=false]{hyperref}

\newcommand{\W}{$\mathcal{W}\,$}
\newcommand{\Wp}{$\mathcal{W}+\,$}

\newcommand{\lw}{\linewidth}

\begin{document}
\title{Approximating Optimal Morphing Attacks using Template Inversion}

\author{Laurent Colbois$^{* 1,2}$, Hatef Otroshi Shahreza$^{* 1,3}$, 
			S\'{e}bastien Marcel$^{1,2}$\\
   $^{1}$Idiap Research Institute, Martigny, Switzerland\\
   $^{2}$Universit\'{e} de Lausanne (UNIL), Lausanne, Switzerland\\
   $^{3}$\'{E}cole Polytechnique F\'{e}d\'{e}rale de Lausanne (EPFL), Lausanne, Switzerland  \\
   {\tt\small \{laurent.colbois,hatef.otroshi,sebastien.marcel\}@idiap.ch}
}
\maketitle
\def\thefootnote{*}\footnotetext{Equal contribution}\def\thefootnote{\arabic{footnote}}

\thispagestyle{empty}

\begin{abstract}
Recent works have demonstrated the feasibility of inverting face recognition systems, enabling to recover convincing face images using only their embeddings. We leverage such template inversion models to develop a novel type of deep morphing attack based on inverting a theoretical optimal morph embedding, which is obtained as an average of the face embeddings of source images. We experiment with two variants of this approach: the first one exploits a fully self-contained embedding-to-image inversion model, while the second leverages the synthesis network of a pretrained StyleGAN network for increased morph realism. We generate morphing attacks from several source datasets and study the effectiveness of those attacks against several face recognition networks. We showcase that our method can compete with and regularly beat the previous state of the art for deep-learning based morph generation in terms of effectiveness, both in white-box and black-box attack scenarios, and is additionally much faster to run. We hope this might facilitate the development of large scale deep morph datasets for training detection models.
\end{abstract}

\section{Introduction}
\begin{figure}[t]
    \centering
    \includegraphics[width=\linewidth]{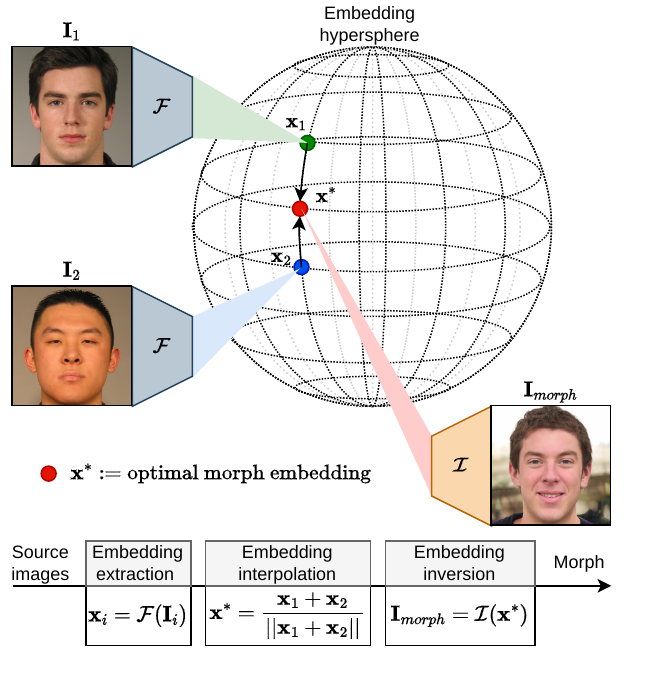}
    \caption{Illustration of the morphing process: face embeddings of the source images are extracted using the FRS $\mathcal{F}$, the corresponding optimal morph embedding is computed by interpolation in the embedding space, then fed back to the template inversion model $\mathcal{I}$ to get the morph. }
    \label{fig:morphing_schema}
\end{figure}

Morphing attacks are a particular type of presentation attack which consists in mixing the faces of two contributing subjects to form a so-called \emph{morph}, and submit it as a reference for enrolment in a face recognition system (FRS), for example as a passport photo. In successful attacks, both contributing subjects can then be authenticated by the FRS while using the same passport, which poses an important security issue. While morphing attacks have historically been generated using landmark-based face editing (LMA), several methods have been proposed over the past years that instead exploit deep learning techniques, in particular Generative Adversarial Networks (GANs). The resulting ``deep'' morphing attacks showcase concerning effectiveness associated with a high realism, although they are generally not yet as successful as LMA ones. Moreover, the most successful GAN-based methods rely on a lengthy latent vector optimization process which renders difficult the generation of a large set of attacks, for example, to create a dataset to train deep morphing attack detection systems.

In parallel of this, recent works in the field of biometric template inversion have demonstrated the feasibility of inverting FRS, i.e., reconstructing a face image starting solely from an extracted face embedding~\cite{Hatef_TI_StyleGAN,template_inversion_icip2022}. This has massive implications, as it enables one to perform any kind of arithmetic operations in the embedding space before going back to the image space. In particular, this method can be exploited for morphing attack generation by first computing interpolations between face embeddings of the source identities, then inverting the interpolated embedding back into the image space.

The aim of this work is to evaluate the effectiveness of using a template inversion process for morph generation. Our main contributions are the following:
\begin{itemize}
    \item{we propose a new strategy to generate approximations of the optimal morph images from the corresponding optimal morph embeddings through template inversion techniques,}
    \item{we introduce two novel deep morphing attack generation algorithms based on two distinct template inversion methods,}
    \item{we evaluate the visual quality of the resulting morphs as well as their attack success rate, both in the settings where the attacked FRS is identical to or different from the inverted FRS.}
\end{itemize}
We observe that the proposed methods are competitive with the previous state of the art in terms of vulnerability, and sometimes even beat it, both in white-box and black-box attack scenarios. Moreover, our morph generation algorithms run orders of magnitude faster than the previous state of the art, making them very practical to generate large deep morphing attack detection datasets.

After contextualizing the state of research on deep morph generation (section \ref{subsec:past_deep_morphs}) and template inversion (section \ref{subsec:realted-works-temp_inv}), we detail our novel morph generation methods in section \ref{sec:morph_generation} and our evaluation process in section \ref{subsec:vulnerability}.
We then discuss the results both qualitatively (section \ref{sec:qualitative_results}) and quantitatively (section \ref{sec:quantitative_results}).

\section{Related works}

\subsection{Deep morph generation}
\label{subsec:past_deep_morphs}
Research on morphing attack generation has originally been focused on \textbf{landmark-based} methods (LMA). Introduced in \cite{ferraraMagicPassport2014a}, those methods proceed by warping the source images to align their facial landmarks, then average pixels between the two warped sources to obtain the morph. As of today, those methods are still typically the most effective at generating morphs able to fool face recognition systems (as evaluated for example in \cite{zhangMIPGANGeneratingStrong2021a}).
More recently, new types of morphing generation techniques have arisen, exploiting recent improvements of deep generative models.
The idea of using a generative adversarial network (GAN) to generate morphs is first introduced in \cite{damerMorGANRecognitionVulnerability2018a}. Their MorGAN model is obtained by jointly training an encoder from the image space to a latent space, and a generator back from the latent space to the image space. Morph latents are then computed by interpolating between the encoded latents of both source images, then fed to the generator to obtain the morph.
\cite{venkateshCanGANGenerated2020} and \cite{sarkarAreGANbasedMorphs2022} expand on this technique by using instead a pretrained StyleGAN2 network \cite{karrasAnalyzingImprovingImage2020}. The image-to-latent encoder is replaced by an optimization process: images are projected in the latent space by finding the latent vector minimizing the perceptual distance between the generated and reference image. Morphs are once again obtained by interpolating the projected latents of the source images and feeding the resulting latent in the generator. With respect to MorGAN, the resulting morphs show significantly improved visual quality, resolution and realism. Several later works take inspiration from those ideas: \cite{damerReGenMorphVisiblyRealistic2021} propose to project LMA morphs in the latent space, before regenerating them with the GAN, in order to get rid of some obvious artifacts. \cite{zhangMorphGANFormerTransformerbasedFace2023} propose similar latent interpolation morphing but replaces the StyleGAN2 backbone by another transformer-based generator architecture.   Finally, \cite{zhangMIPGANGeneratingStrong2021a} get rid of the interpolation step, and instead directly explores the latent space in search of a latent whose associated image is an effective morph. This is done by updating the optimization algorithm to take both source images as input, and by including an additional biometric loss that uses a pretrained FRS.

Other types of generative models have also been used: \cite{blasingameLeveragingDiffusionStrong2023} propose a morphing algorithm based on a diffusion process \cite{hoDenoisingDiffusionProbabilistic2020}. \cite{kellyWorstCaseMorphsTheoretical2022} propose an architecture closer to an autoencoder:  a decoder is trained to reconstruct face images from the combination of their face embedding (extracted using some reference face recognition network), as well as from another latent vector encoding all the face image content \emph{not} related to the identity (this encoder is trained jointly with the decoder). Morphs can then be generated by altering the input face embedding fed to the decoder to use instead a worst-case morph embedding between the two source identities. Conceptually, this approach starts from the same goal as our work (invert an optimal morph embedding back to the image space), but proceeds to it quite differently. Moreover the resulting morphing attack do not show very strong success rates compared to the state-of-the-art. We discuss in section \ref{sec:quantitative_results} in what ways our method differs from theirs.

\subsection{Template inversion}~\label{subsec:realted-works-temp_inv}
Several methods have been proposed in the literature to reconstruct face images from facial templates (embeddings) as template inversion attacks against face recognition systems \cite{zhmoginov2016inverting,cole2017synthesizing,TPAMI2018reconstruction,duong2020vec2face,dong2021towards,vendrow2021realistic,dong2022reconstruct,template_inversion_icip2022,akasaka2022model,ahmad2022inverting}. These methods can be categorized based on the available knowledge from the face feature extractor model into \textit{white-box} and \textit{black-box} methods. In the \textit{white-box} methods, such as \cite{zhmoginov2016inverting,template_inversion_icip2022}, the internal functioning and all the parameters of the face feature extractor model are known, and therefore the feature extractor model is used during training of the face reconstruction network or in gradient-based optimization to reconstruct face images.
In contrast, in the \textit{black-box} methods, such as \cite{TPAMI2018reconstruction,duong2020vec2face,vendrow2021realistic}, the internal functioning of the face feature extractor model is unknown. Therefore, the feature extractor model cannot be used in the training process of the face reconstruction network, but can be used in non-gradient-based optimizations. Since in the \textit{white-box} methods more knowledge of the feature extractor model is available, it is expected (and shown e.g., in \cite{Hatef_TI_StyleGAN}) to achieve better reconstruction performance than \textit{black-box} methods. While most methods are proposed only for either \textit{white-box} or \textit{black-box} scenarios, few methods can be applied to both \textit{white-box} and \textit{black-box} template inversion  \cite{cole2017synthesizing,duong2020vec2face,Hatef_TI_StyleGAN}.

Template inversion methods can be also categorized by their output based on the resolution and the quality of reconstruction. Methods that are based on convolutional neural networks (CNNs), such as \cite{zhmoginov2016inverting,template_inversion_icip2022},  often generate images that suffer from blurriness or other artifacts. 
Whereas, most GAN-based methods generate high-quality and realistic (i.e., \textit{human-face-like}) images. In \cite{duong2020vec2face}, a network based on Pro-GAN~\cite{karras2018progressive} is trained with bijection learning to generate realistic face images. Several other methods use StyleGAN to reconstruct face images from facial templates. 
For instance, in \cite{dong2021towards,Hatef_TI_StyleGAN} StyleGAN is used as the face generator network and the facial templates are mapped to StyleGAN's first or middle layer. Some other works \cite{vendrow2021realistic,dong2022reconstruct} also used optimization on the input of the StyleGAN to find the latent code that can reconstruct the face image.
While the StyleGAN-based methods inherit the leverage of \textit{high-resolution} face generation of StyleGAN, other methods in the literature generate \textit{low-resolution} face images.
Table~\ref{tab:related-works:template-inv} summarizes the template inversion methods proposed in the literature.

\begin{table}[tb]
	\begin{centering}
		\renewcommand{\arraystretch}{1.05}
		\setlength{\tabcolsep}{3pt}
		\caption{Template Inversion methods in the literature.}
		\scalebox{0.85}{
			\begin{tabular}{cccccccc}
				\multirow{2}{*}{\textbf{Ref.}}  &  \textbf{Reconstruction}  &  \textbf{Reconstruction}  &  \textbf{White-box/} & \textbf{Available}\\ 
				& \textbf{Quality}  &  \textbf{Resolution} & \textbf{Black-box} & \textbf{source code}\\
				\toprule
				\cite{zhmoginov2016inverting}&  {low} & {low} & white-box  & \xmark \\\hline
				\cite{cole2017synthesizing} &  {low} & {low} & both  & \xmark \\ \hline
				\cite{TPAMI2018reconstruction}  &  {low} & {low} & black-box  & \cmark \\  \hline
				\cite{duong2020vec2face}  &  {low} & {low} & both  & \xmark \\ \hline
			\cite{template_inversion_icip2022} &  {low} & {low} & white-box  & \cmark \\ \hline
				\cite{akasaka2022model} &  {high} & {low} & black-box  & \xmark \\ \hline
				\cite{ahmad2022inverting} &  {low} & {low} & black-box  & \xmark \\ \hline
				\cite{dong2021towards} &  {high} & {high} & black-box  & \cmark \\  \hline
				\cite{vendrow2021realistic} &  {high} & {high} & black-box  & \cmark \\ \hline
				\cite{dong2022reconstruct} &  {high} & {high} & black-box  & \xmark \\    \hline
				\cite{Hatef_TI_StyleGAN} &  {high} & {high} & both  & \cmark \\  
				\bottomrule
			\end{tabular}\label{tab:related-works:template-inv}
		}
	\end{centering}

\end{table}
 \section{Methodology}

\subsection{Morph generation}
\label{sec:morph_generation}
We introduce a novel method for creating deep morphing attacks, which is grounded in the concept of optimal morph embedding \cite{kellyWorstCaseMorphsTheoretical2022}. The process is illustrated in Figure \ref{fig:morphing_schema}.

Given two source images $\mI_1$ and $\mI_2$, a facial feature extractor $F(.)$ which extract face embeddings $\vx_i := F(\mI) \in \mathcal{X}$, and a distance metric $d(., .)$ on $\mathcal{X}$, the optimal morph embedding is defined by:
\begin{equation}
	\vx^* := \argmin_{\vx \in \mathcal{X}} \left[ d(\vx_1, \vx) + d(\vx_2, \vx)\right] .
\end{equation}
In words, the optimal morph embedding is the face embedding whose biometric distance to the embeddings of both source images is minimized.
If in particular the cosine distance is used as metric, and source embeddings are assumed to be normalized, we have
\begin{equation}\label{eq:optimal-morph}
\vx^* := \frac{\vx_1 + \vx_2}{||\vx_1 + \vx_2||}
\end{equation}
An ideal morphing algorithm would only produce face images whose embedding (for each pair of source images) exactly matches the optimal one. However, while the optimal embedding can be computed, it is in principle only a theoretical construct, and transforming it back into the image space is a priori non trivial. Nevertheless, given recent progress in template inversion methods, we believe this transformation is actually feasible and that it will generate a good approximation of optimal morph images. 

 Our main idea is thus to leverage a biometric template inversion method which will be fed with optimal morph embeddings. With $\mathcal{I}(.)$, a template inversion model trained to invert $\mathcal{F}$, we compute the morph image $\mI_{morph}$ from the optimal embedding as follows:
 \begin{equation}
\mI_{morph} := \mathcal{I}(\vx^*)
\end{equation}We hypothesize that the resulting images are strong candidates for highly effective morphing.
We experiment with two different template inversion systems. The first one (\textbf{base inversion}) consists of a self-contained decoder going from the face embedding space back to the image space, which is expected to be very accurate but also produce images of limited quality and resolution (which is illustrated in section \ref{sec:qualitative_results}). We thus also experiment with a second inversion system (\textbf{GAN-inversion}) which instead learns a mapping from the face embedding space into the latent space of a pretrained StyleGAN model. In doing so, we can leverage the high resolution and realism of StyleGAN generated images, at the possible cost of a lower inversion accuracy. We argue that both those approaches can have their merit depending on whether the \emph{main} focus is to fool the FRS, or to fool some human operator, which is why we choose to experiment with both methods.
The template inversion methods are described in more details in the following section.

\subsubsection{Template inversion}
To reconstruct the morph images from the optimal morph embeddings, we use state-of-the-art white-box template inversion methods proposed in \cite{template_inversion_icip2022} (for low-resolution morph generation) and in \cite{Hatef_TI_StyleGAN} (for high-resolution morph generation). 
Using a white-box template inversion is particularly desirable in our problem of morph generation because we initially have two face images and extract their embeddings with a feature extractor model. Therefore, it is reasonable to consider a \textit{white-box} template inversion method and  use a feature extractor that we have \textit{white-box} knowledge of.

To train the template inversion models, as a preprocessing step, we first normalize the facial templates to have them lie on the embeddings hypersphere (as in Eq.~\ref{eq:optimal-morph}), and then train the template inversion network.
As mentioned in Section~\ref{subsec:realted-works-temp_inv}, the method in \cite{template_inversion_icip2022} generates $112\times112$ \textit{low-resolution} face images in a \textit{white-box} template inversion and the method in 
\cite{Hatef_TI_StyleGAN} generates $1024\times1024$ \textit{high-resolution} and \textit{realistic} face images\footnote{It worth mentioning that the template inversion method proposed in the \cite{Hatef_TI_StyleGAN} is the only method that can generate \textit{high-resolution} face images in \textit{white-box} scenario.}. However, the generated face images by the method proposed in \cite{template_inversion_icip2022} better preserve the identity and achieve a higher attack success rate than the generated face images by the method proposed in \cite{Hatef_TI_StyleGAN} in the reported vulnerability evaluation of the same face recognition systems against template inversion attacks. 

To train the high-resolution template inversion method based on \cite{Hatef_TI_StyleGAN}, we use the exact same GAN training proposed in the original work. 
However, to train the low-resolution template inversion method based on \cite{template_inversion_icip2022}, we update the original method to improve the reconstruction quality. Firstly, we applying an additional perceptual loss function :
\begin{equation}\label{eq:loss_perc}
	\mathcal{L}_\text{Perc}(\mathbf{\hat{I}},\mathbf{I})= ||P(\mathbf{\hat{I}})-P(\mathbf{I})||_1, 
\end{equation}
where $\mathbf{I}$ and $\mathbf{\hat{I}}$ are the original and reconstructed face images, respectively, and $P$ denotes a pre-trained VGG-16~\cite{simonyan2014very} network. Secondly, we also add a skip connection on the convolution blocks.

\subsection{Vulnerability evaluation}
\label{subsec:vulnerability}
To study the effectiveness of our approach, we simulate morphing attacks on several FRS and evaluate the attack success rate. We compare it to previous state-of-the-art methods for deep morph generation, mainly StyleGAN interpolation in both the \W space (\textbf{SG-W}, as in \cite{sarkarAreGANbasedMorphs2022}) and in the \Wp space (\textbf{SG-W+}, as in \cite{venkateshCanGANGenerated2020}), as well as the \textbf{MIPGAN} method \cite{zhangMIPGANGeneratingStrong2021a}. We regenerate StyleGAN interpolation morphs using publicly available tools\footnote{\url{https://gitlab.idiap.ch/bob/bob.morph.sg2}}. For MIPGAN morphs, we reuse the code of the original papers that has gracefully been shared with us by the authors.

The evaluation is decomposed in the generation of morphing attacks from a list of images pairs from a source dataset, followed by the actual vulnerability study where the morphing attacks are enrolled into a biometric system then compared against bona fide probes of the contributing subjects. Following the FRONTEX guideline \cite{frontexBestPracticeTechnical2015}, we calibrate the operating threshold to achieve a FMR of 0.1\% on a reference bona fide protocol. 
We then run the vulnerability protocol evaluation (protocol where the morph are enrolled in the system) and report the Mated Morph Presentation Match Rate as introduced in \cite{scherhagBiometricSystemsMorphing2017}, specifically the MinMax-MMPMR and ProdAvg-MMPMR generalizations, at the operating threshold.

We experiment with two sources datasets commonly used in morphing literature, the Face Research Lab London dataset (FRLL) \cite{debruineFaceResearchLab2017a} and the Face Recognition Grand Challenge (FRGC) \cite{phillipsOverviewFaceRecognition2005}. For FRLL, we select the same morphing pairs as in the AMSL dataset (a morphing dataset also based on FRLL and the morphing method introduced in \cite{neubertExtendedStirTraceBenchmarking2018}). For the vulnerability evaluation, we probe the system with all available frontal poses of the contributing subjects. When working with FRGC, we reuse both the morphing pairs and the probes from \cite{zhangMIPGANGeneratingStrong2021a}. We note in particular that as part of developing the MIPGAN system, a StyleGAN instance trained on FFHQ (\cite{karrasStyleBasedGeneratorArchitecture2019}) has to be fine-tuned on the dataset of contributing subjects. We reuse a MIPGAN system that has been geared towards FRGC morphs, meaning we are not able to also generate FRLL morphs with it. The MIPGAN method will thus only appear in our vulnerability study that uses FRGC as the source dataset.

We consider two face recognition systems (FRS), ArcFace \cite{ArcFaceAdditiveAngular} and ElasticFace \cite{boutrosElasticFaceElasticMargin2022}. For each of them, we train our two considered white-box template inversion systems (base inversion and GAN-inversion), resulting in 4 different template inversion systems. For this stage, the FFHQ dataset \cite{karrasStyleBasedGeneratorArchitecture2019} is used as training data. The same two FRS are then also used as attack target for the vulnerability evaluation. We note that this enables to evaluate how a inversion-based morphing using an inverter trained on some will perform on a different FRS.

The considered morphing approaches are summarized in Table \ref{tab:morphing_methods}. Source code for the replication of those experiments is publicly available.\footnote{\url{https://gitlab.idiap.ch/bob/bob.paper.ijcb2023_inversion_morphing}}

 \begin{table}[t]
 \centering
 \caption{Considered morphing methods. The inversion and GAN-inversion methods are relative to a specific face recognition system, which will be denoted by either AF (ArcFace) or EF (ElasticFace)}
\label{tab:morphing_methods}
\renewcommand{\arraystretch}{1.05}
\resizebox{\columnwidth}{!}{
\begin{tabular}{@{}p{0.3\lw}p{0.7\lw}@{}}
\toprule
Name & Approach \\
\midrule
Inv-AF/EF & Base inversion of optimal template (FRS-dependent) \\
GAN-Inv-AF/EF & GAN-inversion (\W) of optimal template (FRS-dependent) \\
SG-W\cite{sarkarAreGANbasedMorphs2022} & Proj. and interp.. in StyleGAN2 \W~space \\
SG-W+ \cite{venkateshCanGANGenerated2020} & Proj. and interp. in StyleGAN2 \Wp~space \\
MIPGAN\cite{zhangMIPGANGeneratingStrong2021a} & Optimization in StyleGAN2 \Wp~space \\
\bottomrule
\end{tabular}
}
\end{table}

\section{Results and discussion}
\begin{figure}
    \begin{subfigure}[b]{\columnwidth}
    \includegraphics[width=\columnwidth]{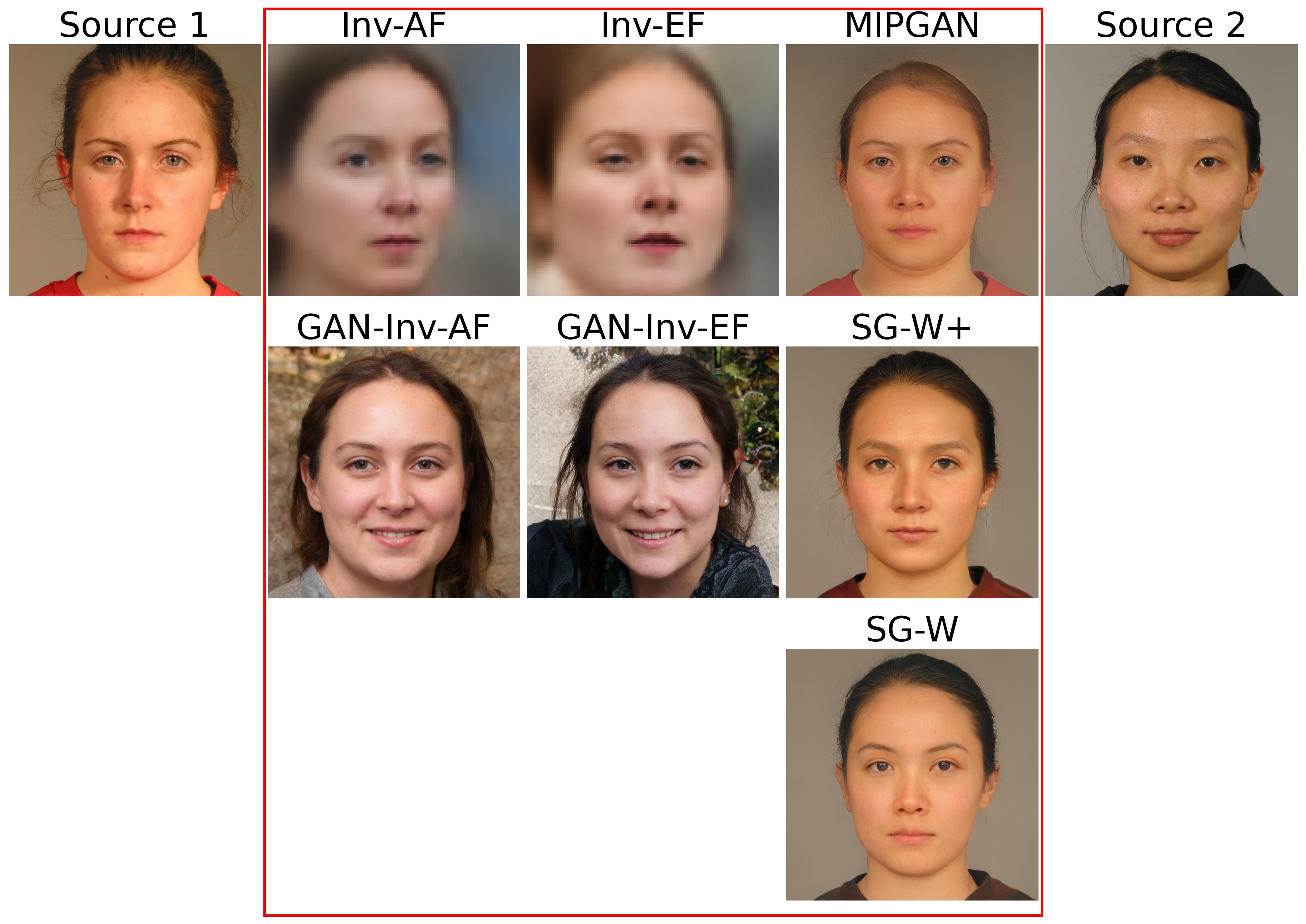}
    \caption{First pair of sources}
    \end{subfigure}
    \begin{subfigure}[b]{\columnwidth}
    \includegraphics[width=\columnwidth]{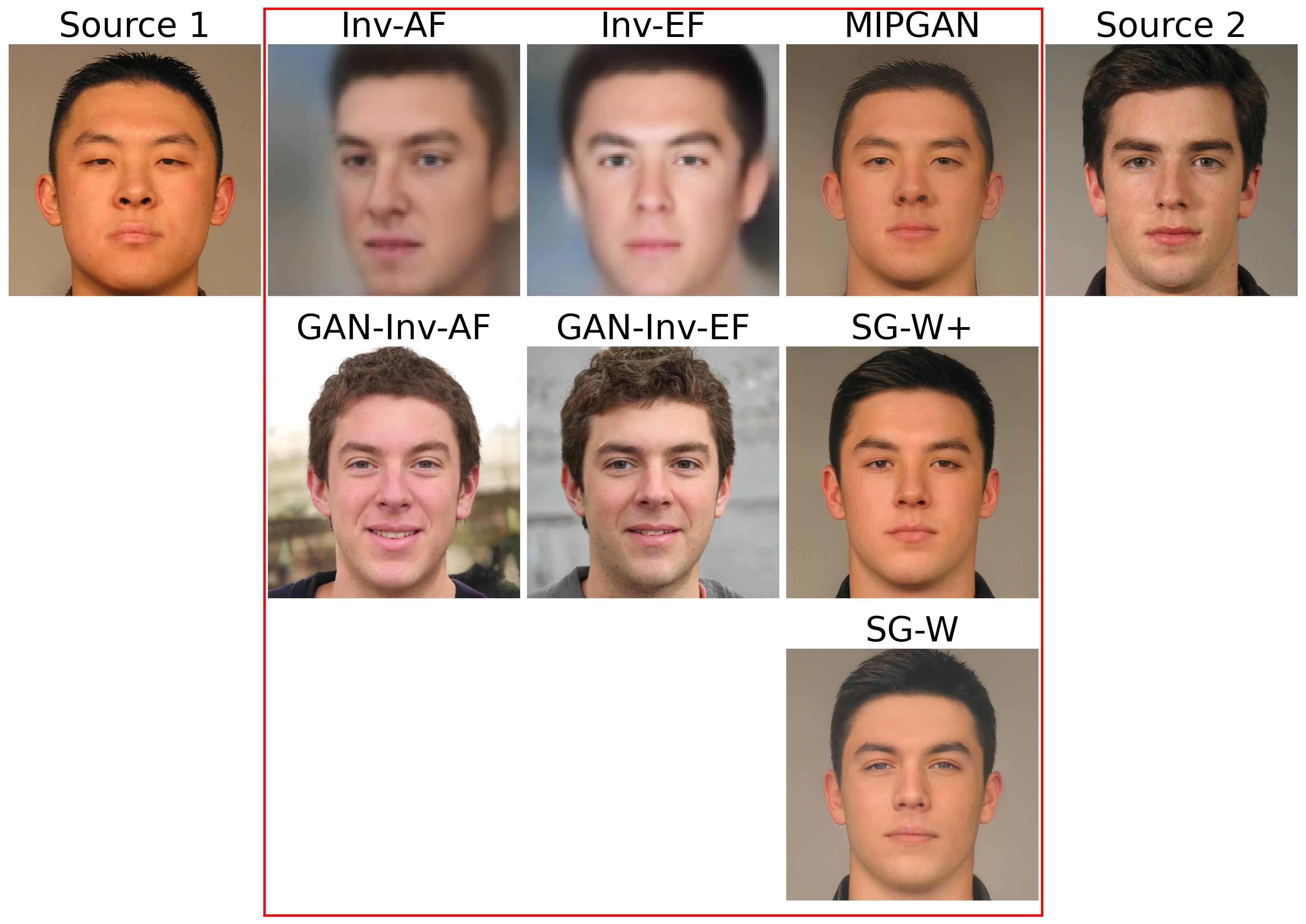}
    \caption{Second pair of sources}
    \end{subfigure}
    \caption{All types of considered deep morphs for two different pairs of source identities.}
    \label{fig:morph_examples}
\end{figure}

\subsection{Qualitative discussion}
\label{sec:qualitative_results}
We start with a discussion of the visual aspects of the obtained morphs which are showcased in Figure \ref{fig:morph_examples}. 
Optimization-based approaches (SG-W, SG-W+ and MIPGAN) typically produce morphs that look realistic, and whose facial features convincingly seem to be mixing elements from both source faces. We note that SG-W morphs are typically less perceptually similar to the sources than SG-W+ and MIPGAN ones. We also note that MIPGAN morphs regularly showcase blurriness artifacts at the border of the head.

Inversion-based approaches, in contrast, generate morphs that look characteristically more distant from the sources. This is not surprising : the inversion model only has access to face embedding data, which ideally should encapsulate solely information crucial for identification, but no other image features. Non-facial properties (e.g., the background) as well as irrelevant face covariates (pose, expression, hairstyle, etc.) can thus be reconstructed in many different ways and should not be constrained by the content of the source images.

The base inversion method generate morphs that are very blurry outside of the face area, and overall have the worst visual realism of all considered methods. One could argue that those inversion-based morphs are unrealistic to use in a real-world scenario. Indeed, the morph having to be submitted for registration to some passport-issuing authority, it is likely that the image might get processed by a human operator at some point in the process. This operator might easily notice that those morphs are blurry or do not look like a passport photo. While we agree with this remark, we suggest that this problem could be circumvented by further post-processing the resulting morphs to splice the face area (which looks realistic) into one of the source images. This is considered for future work.

In contrast, the GAN-inversion morphs have very high realism, given that they can leverage the richness of StyleGAN's face image distribution modelling. In particular, they do not present easily noticeable artifacts, in contrast to MIPGAN morphs in which some ghostly areas can sometimes be observed around the hair or close to the border of the face. In scenarios where fooling a human operator is crucial for the success of the attack, GAN-inversion morphs might thus be already effective with little to no post-processing.

\subsection{Quantitative discussion}
\label{sec:quantitative_results}
Tables \ref{tab:frgc-calib_frgc_mmpmr} and \ref{tab:frgc-calib_frll_mmpmr} present the vulnerability evaluation results, with an operating threshold picked on the \textit{Experiment 2} protocol of the FRGC dataset \cite{phillipsOverviewFaceRecognition2005}. 

We observe that both inversion methods have a significant attack success rate. In particular, when the evaluated FRS is the same that has been inverted to generate the morph, we observe MMPMR in the 40\%-50\% range for GAN-inversion, and even above 90\% for the base inversion method.
However, this corresponds to an unrealistic scenario where the attacked network is fully accessible to generate the attack (\emph{white-box attack}\footnote{We emphasize the distinction between white-box attacks (the attacked FR network is unknown at morph generation time) and white-box template inversion (the inverted FR network is fully accessible to train the inverter). We are here using only white-box inversion methods, but applying them for both white- and black-box morphing attacks.}). In a real-world setting, it is likely that the attacked FRS would be private, and one could fear that inverted morphs obtained by targeting a specific FRS would only be effective against this same FRS. However we observe that this is not the case : when the evaluated FRS is different from the inverted one (\emph{black-box attack}), we do observe a general decrease in the attack success rate, but it still remains at a concerning level. We also observe that there is some trade-off between realism and success rate : while GAN-inversion morphs are of very high quality, they do not perform as well overall as base inversion ones. In some sense, base inversion could be seen as a method that is biased towards fooling the FRS, at the cost of not fooling humans as much, while GAN-inversion is biased in the opposite direction. We argue that depending on the specific details of the enrolment process (in particular how much it is automated or human-processed), both type of attacks might be relevant.

If we compare inversion-based approaches to previous methods, we observe that our base inversion method is competitive with the state of the art.
Indeed, starting with the \emph{white-box} attack scenario, we observe that the Inv-AF systems beats MIPGAN to attack the ArcFace model by a significant margin. We note that MIPGAN should indeed be considered as a white-box attack on ArcFace, given that it uses ArcFace to compute a biometric loss during its own morph generation process. 
In the \emph{black-box} attack scenario, we observe that the Inv-EF system, in particular, showcases an impressive generalization capability. Indeed, when attacking the ArcFace model, the Inv-EF actually beats MIPGAN, \emph{despite MIPGAN having access to the ArcFace system at morph generation time}.
The Inv-AF system also showcases strong black-box attack effectiveness, although it is still beaten by MIPGAN when attacking ElasticFace, but only by a very small margin. Moreover, both base inversion systems always beat SG-W and SG-W+ approaches in the black-box attack scenario, by a wide margin.

The GAN-inversion method, however, is not as effective as previous morphing methods, even though it still showcases concerning attack success rates. But we want to emphasize that this method is learning a mapping from the face embedding space into the \W space  of the used StyleGAN network, which has a limited capacity. As illustration, we see for example that switching from \W to \Wp with the StyleGAN morphing methods drastically improves the attack success rate. MIPGAN is also finding the morph by exploring the \Wp space. We hypothesize that the performance of our GAN-inversion system might be partially limited by this restriction to the \W space, and that learning a new encoder of face embeddings into the \Wp space might give significant returns. However, the process is trickier to train given the high dimensionality of the output space. This is left for future work.

\begin{table}
    \centering
        \caption{MMPMR on the FRGC vulnerability protocol. Threshold is set for FMR@1e-3 on the FRGC Experiment 2 protocol. The FRS column indicates which face recognition system is used at \emph{evaluation} time. We distinguish between white-box ($\square$) and black-box ($\blacksquare$) attacks.}
    \label{tab:frgc-calib_frgc_mmpmr}
    \renewcommand{\arraystretch}{1.05}
    \resizebox{\columnwidth}{!}{
    \begin{tabular}{@{}p{0.1\lw}p{0.3\lw}p{0.3\lw}p{0.3\lw}@{}}
\toprule
FRS & Attack & MinMax-MMPMR (\%) & ProdAvg-MMPMR (\%) \\
\midrule
\multirow[t]{8}{*}{AF}  &$\square$ MIPGAN  & 73.22 & 54.77 \\
& $\square$ Inv-AF  & \bfseries 89.88 & \bfseries 74.76 \\
 & $\square$  GAN-Inv-AF & 42.76 & 22.81 \\
 \cmidrule{2-4}
& $\blacksquare$ SG-W  & 4.32 & 1.44 \\
 & $\blacksquare$ SG-W+  & 60.10 & 39.97 \\
 & $\blacksquare$ Inv-EF  & \bfseries 79.65 & \bfseries  61.46 \\
 & $\blacksquare$ GAN-Inv-EF  & 16.46 & 5.85 \\
\cmidrule{1-4}
\multirow[t]{8}{*}{EF}  &$\square$  Inv-EF & \bfseries 87.78 & \bfseries 74.58 \\
 & $\square$ GAN-Inv-EF & 28.88 & 14.52 \\
 \cmidrule{2-4}
& $\blacksquare$ SG-W & 10.19 & 3.56 \\
 & $\blacksquare$ SG-W+ & 67.63 & 48.90 \\
 & $\blacksquare$ MIPGAN & \bfseries 75.80 & \bfseries 60.10 \\
 & $\blacksquare$ Inv-AF & 75.09 & 58.25 \\

 & $\blacksquare$ GAN-Inv-AF & 28.20 & 14.73 \\
\bottomrule
\end{tabular}

    }
\end{table}
\begin{table}
    \centering
        \caption{MMPMR on the FRLL vulnerability protocol. Threshold is set for FMR@1e-3 on the FRGC Experiment 2 protocol. The FRS column indicates which face recognition system is used at \emph{evaluation} time. We distinguish between white-box ($\square$) and black-box ($\blacksquare$) attacks.}
    \label{tab:frgc-calib_frll_mmpmr}
\renewcommand{\arraystretch}{1.05}
    \resizebox{\columnwidth}{!}{
        \begin{tabular}{@{}p{0.1\lw}p{0.3\lw}p{0.3\lw}p{0.3\lw}@{}}
\toprule
 FRS & Attack &  MinMax-MMPMR (\%) & ProdAvg-MMPMR (\%) \\
\midrule
\multirow[t]{7}{*}{AF} & $\square$ Inv-AF  & \bfseries 97.54 & \bfseries 94.47 \\ 
 & $\square$ GAN-Inv-AF  & 51.58 & 42.39 \\
 \cmidrule{2-4}
 & $\blacksquare$ SG-W  & 1.05 & 0.64 \\
 & $\blacksquare$ SG-W+  & 62.63 & 53.71 \\
 & $\blacksquare$ Inv-EF  & \bfseries 90.70 & \bfseries  85.57 \\
 & $\blacksquare$ GAN-Inv-EF  & 17.19 & 11.93 \\
\cmidrule{1-4}
\multirow[t]{7}{*}{EF}  & $\square$ Inv-EF  & \bfseries 96.67 & \bfseries 93.00 \\
 & $\square$ GAN-Inv-EF  & 37.46 & 29.10 \\
 \cmidrule{2-4}
& $\blacksquare$ SG-W  & 3.07 & 2.06 \\
 & $\blacksquare$ SG-W+  & 72.28 & 62.81 \\
 & $\blacksquare$ Inv-AF  & \bfseries 91.75 & \bfseries 86.80 \\
 & $\blacksquare$ GAN-Inv-AF  & 41.93 & 33.07 \\
\bottomrule
\end{tabular}

    }
\end{table}

We also want to discuss how our method compared by the one proposed in \cite{kellyWorstCaseMorphsTheoretical2022}. This work also attempts to invert an optimal morph embedding by (we stick to the formalism of section \ref{sec:morph_generation})
\begin{enumerate}
\item{complementing the facial feature extractor with a second encoder $E$ trained to extract image features \emph{not} related to the identity,}
\item{training a decoder $D$ to reconstruct an image from $E(\mI_1)$ and $\vx^*$, which should perceptually look like $\mI_1$, but have a face embedding close to $\vx^*$.}
\end{enumerate}
In some sense, this process learns to introduce \emph{imperceptible} signal guided by $\mI_2$ onto $\mI_1$, to bring its face embedding much closer to $\vx^*$.
While effective, this method's main drawback is its reliance on the image features encoder $E$. As this encoder in particular learns general properties of the image distribution of the training set (e.g. color or textures), it might struggle with generalizing to other source datasets (which is unfortunately not evaluated in the original paper), which could show a different color or textures distribution. In contrast, we showcased the reliance on this image features encoder is actually superfluous, and that effective inverted morphs can be obtained solely using optimal morph embeddings as input.

Finally, inversion-based morphing has additional advantages on top of its high attack effectiveness.
Firstly, previous methods rely on a time consuming optimization process exploring the latent space of StyleGAN in order to find either a good projection of the source images (SG-W, SG-W+) or directly a candidate latent that generates an effective morph (MIPGAN). In contrast, once the inversion model is trained, generating inverted morphs is a straightforward process that only requires two forwards passes of the face recognition network and one forward pass of the template inverter. For this reason, morphs can be generated at a speed orders of magnitude faster.
This is showcased in Table \ref{tab:runtime} which presents typical runtimes for end-to-end generation of a single morph. We observe that any of the inversion-based approaches leads to a speed up of around 50x - 75x with respect to MIPGAN. This major speed up could greatly facilitate the creation of large deep morph datasets; for example, to enable the training of effective detection models. 
Secondly, the MIPGAN model is relying on a fine-tuning of a pretrained StyleGAN-FFHQ model using the source dataset. It is yet unclear how much the resulting morphing system is sensitive to the similarity of the source images to the fine-tuning dataset, and it might not always be simple to assemble a new adapted fine-tuning dataset to recalibrate the system for source images that are out-of-distribution.  
We showcased that inversion-based morphing does not suffer from such limitation, as it only relies on the FFHQ dataset at training, but can then be used out-of-the-box on various source data (here FRLL and FRGC) while showing similar success.

\begin{table}
    \centering
        \caption{End-to-end runtime of each generation algorithm to create 1 morph. The measurements are averaged over 10 morph generations.}
    \label{tab:runtime}

    \scalebox{0.85}{
    \begin{tabular}{@{}lr@{}}
\toprule
Attack & Runtime [s] \\
\midrule
SG-W & $372.08\pm1.46$ \\
SG-W+ & $373.67 \pm 2.77$ \\
MIPGAN &
$47.43\pm1.64$\\
Inv-AF & $0.88 \pm 0.05$ \\
Inv-EF & $0.64 \pm 0.01$ \\
GAN-Inv-AF & $0.99 \pm 0.05$ \\
GAN-Inv-EF & $0.75 \pm 0.01$ \\
\bottomrule
\end{tabular}
    }
\end{table}

\subsection{Remark on the detectability of inversion-based morphs}
As we introduce this new deep morphing attack generation method which shows high effectiveness, we are concerned whether it could also challenge common morphing attack detection systems.
We believe that the new proposed method should \emph{not} be drastically more difficult to detect as previous ones. GAN-inversion morphs in particular are still in the end a particular output of a StyleGAN model. StyleGAN images can be reliably detected as showcased in \cite{wangCNNGeneratedImagesAre2020}, even when not sampled in a straightforward manner : \cite{colbois_biosig} showcases that SG-W+ and MIPGAN morphs can also be reliably detected. For morphs obtained with our base inversion method, we do not have as many guarantees; however \cite{wangCNNGeneratedImagesAre2020} claims that the main salient fingerprint of image generative models is caused by upsampling artifacts in the convolutional architecture, a signal that should thus also be present in our inverted morphs given the architecture of the inverter.
To verify this, we propose to actually run our morphs through the detection model from \cite{wangCNNGeneratedImagesAre2020}. It is a GAN-image detector that showcases strong generalization to unknown generators. We run through this detector morphs sets derived from FRGC with all of our considered morphing methods, from which we get a distribution of attack scores. We use a subset of 1000 images of the FFHQ dataset \cite{karrasStyleBasedGeneratorArchitecture2019} to generate a set of bonafide scores. We then report in table \ref{tab:detection} the AUC as well as the equal error rate of the detector using those two sets as respectively positive and negative examples. This corroborates our hypothesis that both base inversion and GAN-inversion morphs can still be detected with reasonable accuracy, but still not as well as previous methods. Improvements in the robustness of existing detectors to new types of morphing attacks is thus still needed. Works in this line of research would benefit from having access to deep morphs datasets showcasing a wide variety of attacks, however such datasets are still scarcely available. We hope our work can contribute to mitigating this scarcity.
\begin{table}
    \centering
    \caption{Performance of the detector from [31] on morphing sets derived from FRGC. 1000 images of FFHQ form the bonafide set.}
    \label{tab:detection}
    \scalebox{0.85}{
    \begin{tabular}{@{}p{0.4\linewidth}p{0.2\linewidth}p{0.2\linewidth}@{}}
    \toprule
    Attack & AUC & EER (\%) \\
    \midrule
    Inv-AF & 0.974 & 7.30 \\
    Inv-EF & 0.948 & 11.61 \\
    GAN-Inv-AF & 0.977 & 6.70 \\
    GAN-Inv-EF & 0.982 & 5.62 \\
    SG-W+ & 1.000 & 0.80 \\
    MIPGAN & 1.000 & 0.18 \\
    \bottomrule
\end{tabular}

    }
\end{table}

\section{Conclusion}
We have demonstrated the feasibility of generating morphing attacks by leveraging template inversion systems to invert optimal morph embeddings.
Both our methods significantly improve the generation speed with respect to the previous state of the art. Moreover, our base inversion morphing method is competitive with the previous state-of-the-art in terms of attack success rate, and often beats it by a large margin, both in the white-box and black-box attack scenario. The Inv-EF system in particular showcases such strong generalization that even in a black-box attack scenario (when attacking the ArcFace model), it is still more effective that MIPGAN, despite the latter actually using ArcFace to compute a biometric loss as part of the morph generation process. 

One main limitation of this base inversion method is that the resulting morphs are somewhat lacking in realism. We believe that a further post-processing of those morphs to splice them back into one of the source images could mitigate the visual realism issue while not losing too much in attack success rate. 
Our GAN-inversion morphing method does display great realism, but generates attacks with lower (but still problematic) effectiveness.
We hypothesize that this latter method could be improved by mapping face embeddings into the \Wp space of StyleGAN (which has greater capacity) instead of the \W space as is done currently. Indeed, we note that with StyleGAN interpolation methods for example, the simple switch from \W to \Wp drastically improve the effectiveness of the attack.
Interestingly, despite their reliance on a white-box access to some FRS, the inversion morphing attacks stays successful when used to attack some other unseen FRS. There is still however a decrease in effectiveness in this latter case; improving the generalization to unseen FRS is another direction that might be interesting for future work.

Finally, our methods enable fast generation of large scale morphing datasets, which we hope could facilitate the development and training of deep morphing attack detection systems.

\iftoggle{cvprfinal}{
\section*{Acknowledgments}
This research is based upon work supported by the H2020 TReSPAsS-ETN Marie Sk\l{}odowska-Curie early training network (grant agreement 860813). 
This work was also supported by the Swiss Center for Biometrics Research \& Testing and the Idiap Research Institute. 
}{}

{\small
\bibliographystyle{ieee}
\bibliography{egbib}
}

\end{document}